\documentclass[10pt,twocolumn,letterpaper]{article}

 \usepackage{iccv}              %

\definecolor{iccvblue}{rgb}{0.21,0.49,0.74}
\usepackage[pagebackref,breaklinks,colorlinks,allcolors=iccvblue]{hyperref}

\def\paperID{*****} %
\def\confName{ICCV}
\def\confYear{2025}

\title{Better, But Not Sufficient: Testing Video ANNs Against Macaque IT Dynamics}

\author{Matteo Dunnhofer\\
York University, Toronto, Canada\\
University of Udine, Italy\\
{\tt\small matteo24@yorku.ca}
\and
Christian Micheloni\\
University of Udine, Italy \\
{\tt\small christian.micheloni@uniud.it}
\and
Kohitij Kar\\
York University, Toronto, Canada\\
{\tt\small k0h1t1j@yorku.ca}
}

\begin{document}
\maketitle
\begin{abstract}
Feedforward artificial neural networks (ANNs) trained on static images remain the dominant models of the the primate ventral visual stream, yet they are intrinsically limited to static computations. The primate world is dynamic, and the macaque ventral visual pathways, specifically the inferior temporal (IT) cortex not only supports object recognition but also encodes object motion velocity during naturalistic video viewing. Does IT’s temporal responses reflect nothing more than time-unfolded feedforward transformations, framewise features with shallow temporal pooling, or do they embody richer dynamic computations? We tested this by comparing macaque IT responses during naturalistic videos against static, recurrent, and video-based ANN models. Video models provided modest improvements in neural predictivity, particularly at later response stages, raising the question of what kind of dynamics they capture. To probe this, we applied a stress test: decoders trained on naturalistic videos were evaluated on “appearance-free” variants that preserve motion but remove shape and texture. IT population activity generalized across this manipulation, but all ANN classes failed. Thus, current video models better capture appearance-bound dynamics rather than the appearance-invariant temporal computations expressed in IT, underscoring the need for new objectives that encode biological temporal statistics and invariances.
\end{abstract}
    
\section{Introduction}
\label{sec:intro}

Over the past decade, feedforward artificial neural networks (ANNs) \cite{krizhevsky2012imagenet,he2016deep} trained on large image datasets \cite{deng2009imagenet} have become the dominant models of the primate ventral visual stream \cite{kar2024quest}. These networks approximate responses in areas such as V4 \cite{bashivan2019neural} and inferior temporal (IT) cortex \cite{yamins2014performance,schrimpf2018brain, khaligh2014deep}, and their internal representations support linear readouts of object category and position that match behavioral performance on recognition tasks \cite{rajalingham2018large,schrimpf2018brain}. This success has reinforced the idea that feedforward image computation may be a sufficient account of core object vision. Yet the biological ventral stream does not operate on static images—it is embedded in a dynamic world in which objects move, change pose, and interact with one another.

A growing body of evidence shows that macaque IT cortex supports computations that extend beyond recognition of static form \cite{kar2019evidence}. In particular, IT neurons encode object motion and velocity, suggesting that ventral cortex contributes directly to dynamic vision \cite{ramezanpour2024object}. These findings raise a critical scientific gap: if feedforward ANNs trained on static images explain IT responses to static objects, what explains IT’s ability to process naturalistic videos? Is IT’s temporal structure merely a “time-unfolded” extension of feedforward computations -- i.e. framewise features linked together by shallow temporal pooling -- or does it instead reflect richer dynamic mechanisms that current models fail to capture?

Recent advances in ANN modeling have introduced recurrence and video training regimes that, at least superficially, bring models closer to biological reality \cite{kubilius2019brain,nayebi2022recurrent}. Recurrent networks recycle feature activations across time, and video ANNs trained for tasks such as action recognition \cite{feichtenhofer2019slowfast,carreira2017quo,bertasius2021space,li2024videomamba} or object tracking \cite{bertinetto2016fully,ravi2024sam,dunnhofer2023visual} can extract temporal regularities across frames. In our study, some of these models have shown modest improvements in predicting IT activity during video viewing. However, it remains unclear what kind of temporal signals these models capture. Are they sensitive primarily to appearance-bound transients such as evolving texture or pose? Or do they approximate the appearance-invariant motion computations expressed in IT? This is the smaller but crucial gap our work addresses.

We directly compared macaque IT responses during short naturalistic videos (300 ms; 18 frames) against multiple classes of ANN models, including static feedforward, recurrent, and video-based architectures. We assessed both frame-level predictivity and video-level decoding to test whether models capture the evolving temporal responses observed in IT.

Video ANNs provided modest improvements over static and recurrent baselines, particularly at later response stages, suggesting that temporal training objectives confer some alignment with IT dynamics. However, these improvements raised the deeper question of what underlying computations they reflect. To probe this, we implemented a stress test inspired by recent neurophysiological observations \cite{ilic2022appearance,ramezanpour2024object}: decoders trained on naturalistic videos were evaluated on “appearance-free” variants that preserved object motion trajectories but removed texture and shape cues.

Macaque IT population activity generalized robustly across this manipulation. Motion direction and velocity remained decodable from IT even when appearance information was stripped away. In stark contrast, all ANN classes -- including video ANNs -- failed under this stress test. While these models could track motion in naturalistic clips, their representations collapsed when only motion remained. Moreover, neither time-unfolded feedforward sequences nor simple recurrent features reproduced the temporal response structure observed in IT, indicating that current architectures lack the necessary invariances.

Our results reveal a layered picture. On one hand, video training does provide modest gains in aligning ANN features with late IT responses, highlighting progress toward dynamic modeling. On the other, the failure on appearance-free videos exposes a fundamental limitation: current models capture appearance-bound temporal signals but miss the appearance-invariant motion computations supported by IT. These findings argue that IT dynamics cannot be approximated as a simple extension of feedforward computation. Building biologically aligned video models will require new objectives and architectural constraints that encode temporal statistics and invariances observed in IT populations, moving beyond static supervision and shallow temporal pooling.

\begin{figure*}[t]
    \centering
    \includegraphics[width=1\linewidth]{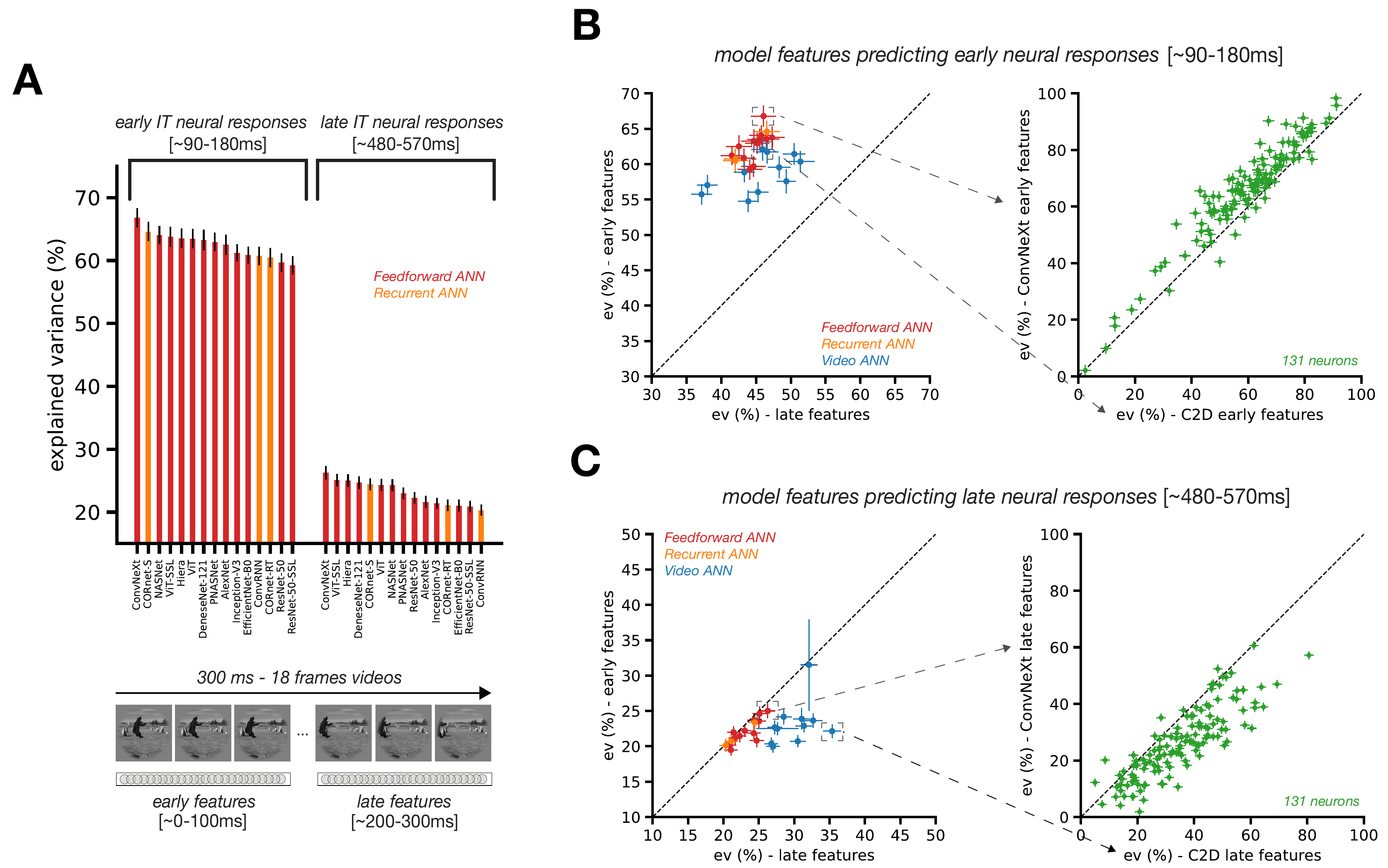}
    \caption{\textbf{A}. Neural predictivity of early and late features extracted from early and late video frames by static feed-forward and recurrent ANNs. We report mean and standard error across neurons. \textbf{B}. (Left) Neural predictivity of early and late static and video ANNs on early neural responses. We report mean and standard error across neurons. (Right) Per-neuron explained variance by the best static and video model in early responses. We report mean and standard error across repetitions. \textbf{C}. (Left) Neural predictivity of early and late static and video ANNs on late neural responses. We report mean and standard error across neurons. (Right) Per-neuron explained variance by the best static and video model in late responses. We report mean and standard error across repetitions.}
    \label{fig:fig1}
\end{figure*}

\section{Results}

As mentioned above, the key question is whether IT’s temporal responses to naturalistic videos can be reduced to time-unfolded feedforward transformations or whether they reflect richer dynamics. We recorded activity from 131 IT sites while two monkeys passively viewed 300-ms videos under strict fixation control, ensuring purely visual responses. We then compared IT dynamics to features from static feedforward, recurrent, and video-trained ANNs. Analyses proceeded in two stages: first, measuring neural predictivity over time to test for model gains; second, applying a stress test with “appearance-free” videos to distinguish genuine dynamic computations from appearance-bound signals.

\subsection{Comparison of IT predictivity of feedforward and video ANNs during dynamic vision}

We first evaluated whether video-trained ANNs provide a better account of IT responses during naturalistic video viewing than static feedforward models. Figure 1A–C summarizes this comparison. Feedforward models, unfolded in time, were able to explain a significant fraction of IT variance, particularly for early neural responses ($\sim$90–180 ms). However, their predictivity dropped markedly for later IT responses ($\sim$480–570 ms). Video ANNs showed modest but reliable improvements in this late window, suggesting that temporal training objectives confer some benefit in capturing the extended dynamics of IT. Recurrent models performed comparably to feedforward networks, indicating that shallow recurrence alone does not bridge the gap.

Critically, per-neuron analyses (Figure 1B–C) revealed that while some IT neurons were equally well captured by both static and video models, a subset of neurons was better explained by video ANNs during their late response phase and better by static ANNs in their early response phases. This pattern suggests that video training partially aligns model features with later phases of IT dynamics but does not fully account for the heterogeneity of temporal responses across the IT population.

\begin{figure*}[hbt]
    \centering
    \includegraphics[width=1\linewidth]{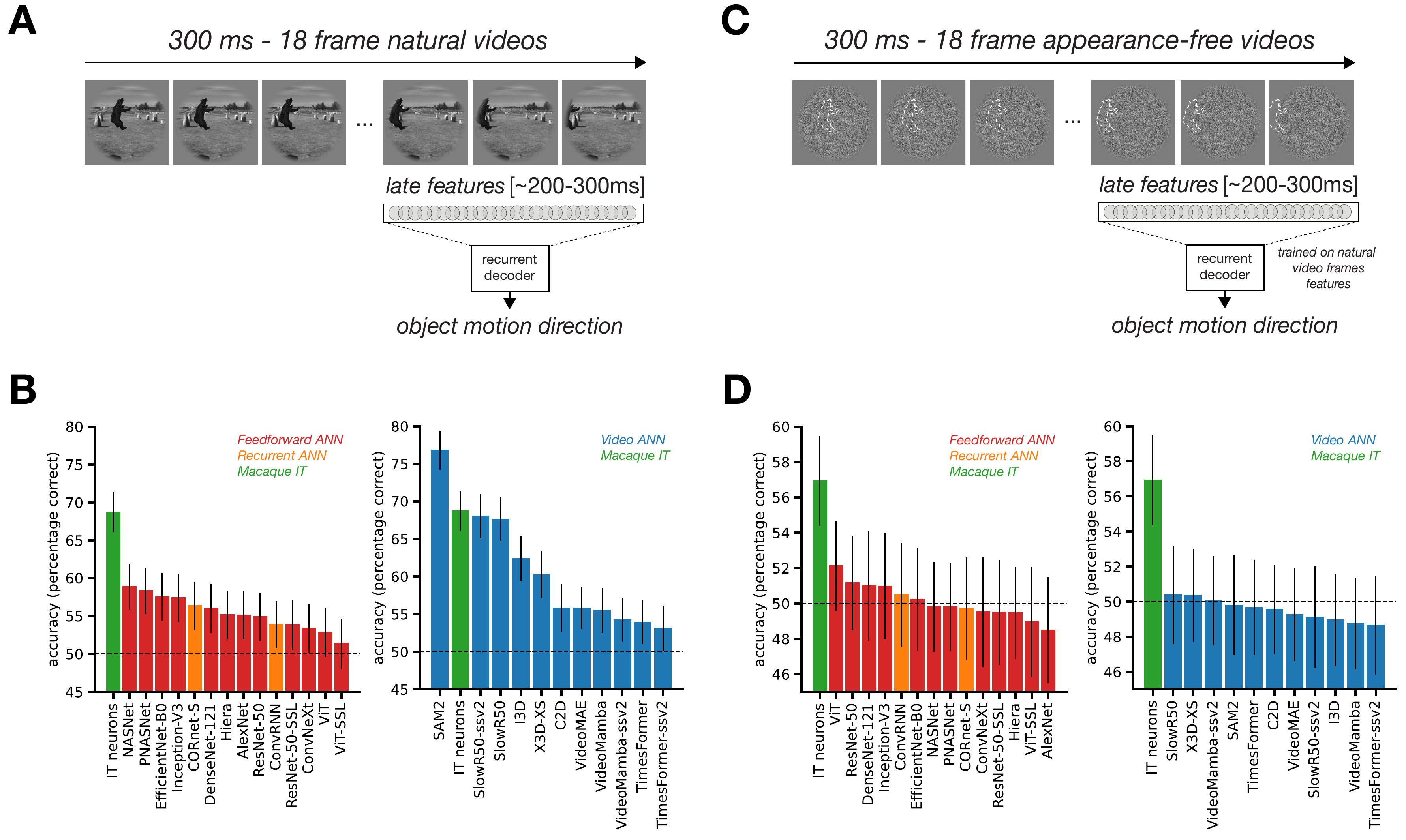}
    \caption{\textbf{A}. LSTM-based decoding mechanism used to evaluate how well the model represents dynamic object properties in natural videos.
    \textbf{B}. Decoding accuracy of object-motion direction from late model features and late IT responses (83 neurons) for static and video ANNs on natural videos. We report mean and standard error across videos.
    \textbf{C}. LSTM-based decoding mechanism used to evaluate how well the model represents dynamic object properties in appearance-free videos. Appearance-free videos consist of frame sequences where appearance is replaced by random pixel values that move according to the original object motion \cite{ilic2022appearance}.
    \textbf{D}. Decoding accuracy of object-motion direction from late model features and late IT responses (83 neurons) for static and video ANNs on appearance-free videos. We report mean and standard error across videos. }
    \label{fig:fig2}
\end{figure*}

\subsection{Video ANNs and not macaque IT fail to discriminate motion direction without object-appearance}

As expected from their improved late-phase predictivity (Section 2.1), video ANNs were more accurate than feedforward and recurrent networks in supporting motion-direction decoding when tested on naturalistic videos (Figure 2B). 
This pattern mirrors macaque IT, where late-phase responses also carried stronger motion information than early responses \cite{ramezanpour2024object}.
Thus, temporal training appears to confer some alignment between ANN and IT dynamics.

However, the critical stress test revealed a sharp divergence. When decoders trained on naturalistic videos were evaluated on appearance-free variants (similar to \cite{ilic2022appearance}), macaque IT maintained robust motion direction decoding accuracies (Figure 2D, left), reflecting an ability to generalize across appearance manipulations. In contrast, video ANNs collapsed to chance performance (Figure 2D, right, IT decoding accuracy vs. model decoding accuracy t-test p-value $<$ 0.001). Despite their gains on naturalistic stimuli, ANN representations were unable to support motion discrimination once appearance cues were stripped away.

This result shows that while IT dynamics encode appearance-invariant motion, current ANN models remain bound to framewise appearance changes, exposing a key limitation of existing video architecture and training.

\section{Discussion}

Our results show that current video ANNs, while modestly improving IT predictivity during naturalistic video viewing, fall short of capturing the full range of temporal computations observed in macaque IT. The late-phase ($\sim$480-570ms) gains offered by video training objectives suggest some sensitivity to temporal statistics, but the catastrophic failure under appearance-free conditions (Figure 2D, right panel) reveals that these models remain tied to framewise appearance cues rather than encoding appearance-invariant motion dynamics. 

Beyond this broad conclusion, we believe future analyses should be designed to identify multiple dynamic properties of IT responses that can serve as quantitative constraints for modeling. First, measuring how quickly neurons change their responses over time to reveal heterogeneity in temporal stability versus volatility across the IT population. Second, quantifing the proportion of neurons whose activity is modulated primarily by object identity versus those that encode both object identity and its dynamics. Third, characterizing the diversity of temporal response profiles, potentially showing that some IT neurons maintain consistent signals while others undergo systematic shifts as videos unfold. For machine learning-based computer vision, this could provide a roadmap. Simply training video ANNs on large action-recognition datasets nudges them toward IT but does not fully reproduce its representational structure. Incorporating empirically measured temporal properties—such as the distribution of neuronal reliability, the prevalence of mixed selectivity for object and motion, and the population-level geometry of temporal responses—into model design or optimization may drive the next generation of biologically aligned architectures. Such constraints could be introduced through auxiliary objectives, architectural biases, or dynamic regularizers that force models to respect temporal invariances expressed in IT.

In sum, we find that ANN models of dynamic vision are close but not enough: they partially capture IT’s late-phase video response dynamics but fail when challenged with appearance-free motion. At the same time, our analyses set the basis for a richer set of empirical constraints that can guide the development of new architectures. By grounding future models in both behavioral benchmarks and the dynamic statistics of IT responses, we can move toward a more complete understanding of how the ventral stream supports vision in a dynamic world.

\noindent\paragraph{Acknowledgments.} MD received funding from the European Union’s Horizon Europe research and innovation programme under the Marie Skłodowska-Curie grant agreement n. 101151834 (PRINNEVOT CUP G23C24000910006). KK has been supported by funds from the Canada Research Chair Program, Google Research, the Canada First Research Excellence Funds (VISTA Program), and the National Sciences and Engineering Research Council of Canada (NSERC).

{
    \small
    \bibliographystyle{ieeenat_fullname}
    \bibliography{main}

\begin{thebibliography}{21}
\providecommand{\natexlab}[1]{#1}
\providecommand{\url}[1]{\texttt{#1}}
\expandafter\ifx\csname urlstyle\endcsname\relax
  \providecommand{\doi}[1]{doi: #1}\else
  \providecommand{\doi}{doi: \begingroup \urlstyle{rm}\Url}\fi

\bibitem[Bashivan et~al.(2019)Bashivan, Kar, and DiCarlo]{bashivan2019neural}
Pouya Bashivan, Kohitij Kar, and James~J DiCarlo.
\newblock Neural population control via deep image synthesis.
\newblock \emph{Science}, 364\penalty0 (6439):\penalty0 eaav9436, 2019.

\bibitem[Bertasius et~al.(2021)Bertasius, Wang, and Torresani]{bertasius2021space}
Gedas Bertasius, Heng Wang, and Lorenzo Torresani.
\newblock Is space-time attention all you need for video understanding?
\newblock In \emph{International Conference on Machine Learning}, 2021.

\bibitem[Bertinetto et~al.(2016)Bertinetto, Valmadre, Henriques, Vedaldi, and Torr]{bertinetto2016fully}
Luca Bertinetto, Jack Valmadre, Joao~F Henriques, Andrea Vedaldi, and Philip~HS Torr.
\newblock Fully-convolutional siamese networks for object tracking.
\newblock In \emph{European Conference on Computer Vision}, pages 850--865. Springer, 2016.

\bibitem[Carreira and Zisserman(2017)]{carreira2017quo}
Joao Carreira and Andrew Zisserman.
\newblock Quo vadis, action recognition? a new model and the kinetics dataset.
\newblock In \emph{IEEE Conference on Computer Vision and Pattern Recognition}, pages 6299--6308, 2017.

\bibitem[Deng et~al.(2009)Deng, Dong, Socher, Li, Li, and Fei-Fei]{deng2009imagenet}
Jia Deng, Wei Dong, Richard Socher, Li-Jia Li, Kai Li, and Li Fei-Fei.
\newblock Imagenet: A large-scale hierarchical image database.
\newblock In \emph{IEEE Conference on Computer Vision and Pattern Recognition}, pages 248--255. Ieee, 2009.

\bibitem[Dunnhofer et~al.(2023)Dunnhofer, Furnari, Farinella, and Micheloni]{dunnhofer2023visual}
Matteo Dunnhofer, Antonino Furnari, Giovanni~Maria Farinella, and Christian Micheloni.
\newblock Visual object tracking in first person vision.
\newblock \emph{International Journal of Computer Vision}, 2023.

\bibitem[Feichtenhofer et~al.(2019)Feichtenhofer, Fan, Malik, and He]{feichtenhofer2019slowfast}
Christoph Feichtenhofer, Haoqi Fan, Jitendra Malik, and Kaiming He.
\newblock Slowfast networks for video recognition.
\newblock In \emph{IEEE/CVF International Conference on Computer Vision}, pages 6202--6211, 2019.

\bibitem[He et~al.(2016)He, Zhang, Ren, and Sun]{he2016deep}
Kaiming He, Xiangyu Zhang, Shaoqing Ren, and Jian Sun.
\newblock Deep residual learning for image recognition.
\newblock In \emph{IEEE Conference on Computer Vision and Pattern Recognition}, pages 770--778, 2016.

\bibitem[Ilic et~al.(2022)Ilic, Pock, and Wildes]{ilic2022appearance}
Filip Ilic, Thomas Pock, and Richard~P Wildes.
\newblock Is appearance free action recognition possible?
\newblock In \emph{European Conference on Computer Vision}, pages 156--173. Springer, 2022.

\bibitem[Kar and DiCarlo(2024)]{kar2024quest}
Kohitij Kar and James~J DiCarlo.
\newblock The quest for an integrated set of neural mechanisms underlying object recognition in primates.
\newblock \emph{Annual Review of Vision Science}, 10, 2024.

\bibitem[Kar et~al.(2019)Kar, Kubilius, Schmidt, Issa, and DiCarlo]{kar2019evidence}
Kohitij Kar, Jonas Kubilius, Kailyn Schmidt, Elias~B Issa, and James~J DiCarlo.
\newblock Evidence that recurrent circuits are critical to the ventral stream’s execution of core object recognition behavior.
\newblock \emph{Nature Neuroscience}, 22\penalty0 (6):\penalty0 974--983, 2019.

\bibitem[Khaligh-Razavi and Kriegeskorte(2014)]{khaligh2014deep}
Seyed-Mahdi Khaligh-Razavi and Nikolaus Kriegeskorte.
\newblock Deep supervised, but not unsupervised, models may explain it cortical representation.
\newblock \emph{PLoS Computational Biology}, 10\penalty0 (11):\penalty0 e1003915, 2014.

\bibitem[Krizhevsky et~al.(2012)Krizhevsky, Sutskever, and Hinton]{krizhevsky2012imagenet}
Alex Krizhevsky, Ilya Sutskever, and Geoffrey~E Hinton.
\newblock Imagenet classification with deep convolutional neural networks.
\newblock \emph{Advances in Neural Information Processing Systems}, 25, 2012.

\bibitem[Kubilius et~al.(2019)Kubilius, Schrimpf, Kar, Rajalingham, Hong, Majaj, Issa, Bashivan, Prescott-Roy, Schmidt, et~al.]{kubilius2019brain}
Jonas Kubilius, Martin Schrimpf, Kohitij Kar, Rishi Rajalingham, Ha Hong, Najib Majaj, Elias Issa, Pouya Bashivan, Jonathan Prescott-Roy, Kailyn Schmidt, et~al.
\newblock Brain-like object recognition with high-performing shallow recurrent anns.
\newblock \emph{Advances in Neural Information Processing Systems}, 32, 2019.

\bibitem[Li et~al.(2024)Li, Li, Wang, He, Wang, Wang, and Qiao]{li2024videomamba}
Kunchang Li, Xinhao Li, Yi Wang, Yinan He, Yali Wang, Limin Wang, and Yu Qiao.
\newblock Videomamba: State space model for efficient video understanding.
\newblock In \emph{European Conference on Computer Vision}, pages 237--255. Springer, 2024.

\bibitem[Nayebi et~al.(2022)Nayebi, Sagastuy-Brena, Bear, Kar, Kubilius, Ganguli, Sussillo, DiCarlo, and Yamins]{nayebi2022recurrent}
Aran Nayebi, Javier Sagastuy-Brena, Daniel~M Bear, Kohitij Kar, Jonas Kubilius, Surya Ganguli, David Sussillo, James~J DiCarlo, and Daniel~LK Yamins.
\newblock Recurrent connections in the primate ventral visual stream mediate a trade-off between task performance and network size during core object recognition.
\newblock \emph{Neural Computation}, 34\penalty0 (8):\penalty0 1652--1675, 2022.

\bibitem[Rajalingham et~al.(2018)Rajalingham, Issa, Bashivan, Kar, Schmidt, and DiCarlo]{rajalingham2018large}
Rishi Rajalingham, Elias~B Issa, Pouya Bashivan, Kohitij Kar, Kailyn Schmidt, and James~J DiCarlo.
\newblock Large-scale, high-resolution comparison of the core visual object recognition behavior of humans, monkeys, and state-of-the-art deep artificial neural networks.
\newblock \emph{Journal of Neuroscience}, 38\penalty0 (33):\penalty0 7255--7269, 2018.

\bibitem[Ramezanpour et~al.(2024)Ramezanpour, Ilic, Wildes, and Kar]{ramezanpour2024object}
Hamidreza Ramezanpour, Filip Ilic, Richard~P Wildes, and Kohitij Kar.
\newblock Object motion representation in the macaque ventral stream--a gateway to understanding the brain’s intuitive physics engine.
\newblock \emph{bioRxiv}, pages 2024--02, 2024.

\bibitem[Ravi et~al.(2024)Ravi, Gabeur, Hu, Hu, Ryali, Ma, Khedr, R{\"a}dle, Rolland, Gustafson, et~al.]{ravi2024sam}
Nikhila Ravi, Valentin Gabeur, Yuan-Ting Hu, Ronghang Hu, Chaitanya Ryali, Tengyu Ma, Haitham Khedr, Roman R{\"a}dle, Chloe Rolland, Laura Gustafson, et~al.
\newblock Sam 2: Segment anything in images and videos.
\newblock \emph{arXiv preprint arXiv:2408.00714}, 2024.

\bibitem[Schrimpf et~al.(2018)Schrimpf, Kubilius, Hong, Majaj, Rajalingham, Issa, Kar, Bashivan, Prescott-Roy, Geiger, et~al.]{schrimpf2018brain}
Martin Schrimpf, Jonas Kubilius, Ha Hong, Najib~J Majaj, Rishi Rajalingham, Elias~B Issa, Kohitij Kar, Pouya Bashivan, Jonathan Prescott-Roy, Franziska Geiger, et~al.
\newblock Brain-score: Which artificial neural network for object recognition is most brain-like?
\newblock \emph{BioRxiv}, page 407007, 2018.

\bibitem[Yamins et~al.(2014)Yamins, Hong, Cadieu, Solomon, Seibert, and DiCarlo]{yamins2014performance}
Daniel~LK Yamins, Ha Hong, Charles~F Cadieu, Ethan~A Solomon, Darren Seibert, and James~J DiCarlo.
\newblock Performance-optimized hierarchical models predict neural responses in higher visual cortex.
\newblock \emph{Proceedings of the national academy of sciences}, 111\penalty0 (23):\penalty0 8619--8624, 2014.

\end{thebibliography}
}

\end{document}